\documentclass{article} 
\usepackage[margin=1in]{geometry}
\usepackage{times}
\usepackage{tikz}

\usepackage{hyperref}
\usepackage{url}

\title{Interpretable Climate Change Modeling With Progressive Cascade Networks}

\author{Charles Anderson$^1$ \and Jason Stock$^1$ \and David Anderson$^2$}

\date{%
  \small
    $^1$Colorado State University \texttt{chuck.anderson@colostate.edu}\\%
    $^2$Momentiv.AI\\[2ex]%
    \today
}

\begin{document}

\maketitle

\begin{abstract}
Typical deep learning approaches to modeling high-dimensional data often result in complex models that do not easily reveal a new understanding of the data.  Research in the deep learning field is very actively pursuing new methods to interpret deep neural networks and to reduce their complexity.  An approach is described here that starts with linear models and incrementally adds complexity only as supported by the data. An application is shown in which models that map global temperature and precipitation to years are trained to investigate patterns associated with changes in climate.

\end{abstract}

\section{Introduction}

Deep networks have been used successfully to model many complex
relationships in data from a wide variety of domains.  Both practice
and theory suggest that large, deep networks may generalize better to
untrained data samples. This leads to important questions regarding the
interpretability of such large networks to understand the limitations
and trustworthiness of such models.

In many studies involving measurements of the natural world, domain
experts are most familiar with simple statistical analysis, such as
linear regression.  Therefore, to simplify the interpretability of
models, linear models should be the first step.  This can also result
in a better trust of the modeling effort by the domain experts.

This paper describes such an approach that starts with a simple model
consisting of a single-layer, linear network and
incrementally grows the complexity of the model.  A more complex
network is added only if its inclusion reduces the error on a
validation set. Each network can be analyzed to determine the patterns
in the data that are discovered by the networks to lead to improved
modeling results.

\begin{figure}[!ht]
\centering
  \begin{tabular}{cc}
\resizebox{0.48\textwidth}{!}{%
  \begin{tikzpicture}  
  \tikzset{
      node distance=3cm,
    charge node/.style={inner sep=0pt},
    pics/sum block/.style n args={4}{
      code={
        \path node (n) [draw, circle, inner sep=0pt, minimum size=9mm] {}
          (n.north) +(0,-1.5mm) node [charge node] {$#1$}
          (n.south) +(0,1.5mm) node [charge node] {$#2$}
          (n.west) +(1.5mm,0) node [charge node] {$#3$}
          (n.east) +(-1.5mm,0) node [charge node] {$#4$}
          ;
      }
    }
  }
    \tikzstyle{output layer} = [rectangle, minimum width=1cm, minimum height=2cm, centered, draw=black, fill=blue!60!white]
    \tikzstyle{hidden layer} = [rectangle, minimum width=1cm, minimum height=2cm, centered, draw=black, fill=cyan]
    \tikzstyle{arrow} = [thick,->,>=stealth]
    \node (layer_1_1) [output layer] {};
    
    \node (layer_2_2) [output layer, below of=layer_1_1] {};
    \node (layer_2_1) [hidden layer, left of=layer_2_2] {};
    \draw [arrow] (layer_2_1) -- (layer_2_2);
    
    \node (layer_3_3) [output layer, below of=layer_2_2, yshift=-2cm] {};
    \node (layer_3_2) [hidden layer, left of=layer_3_3] {};
    \node (layer_3_1) [hidden layer, left of=layer_3_2] {};
    \path (layer_3_1) -- (layer_3_2) node (layerdots) [font=\Huge, midway] {$\dots$};
    \draw [thick] (layer_3_1) -- (layerdots);
    \draw [arrow] (layerdots) -- (layer_3_2);
                                                    
    \draw [arrow] (layer_3_2) -- (layer_3_3);
    \path (layer_2_1) -- (layer_3_2) node [font=\Huge, midway, sloped] {$\dots$};

    \node (input) [font=\huge, left of=layer_2_1, xshift=-2cm] {$X_i$};
    \draw[arrow] (input) edge[out=0,in=180,->] (layer_1_1);
    \draw[arrow] (input) edge[out=0,in=180,->] (layer_2_1);
    \draw[arrow] (input) edge[out=0,in=180,->] (layer_3_1);
    
    \pic [local bounding box=err1, right of=layer_1_1] {sum block={+}{}{-}{}};
    \draw[arrow] (layer_1_1) -- (err1);
                                                    
    \pic [local bounding box=err2, right of=layer_2_2] {sum block={+}{}{-}{}};
    \draw[arrow] (layer_2_2) -- (err2);
                                                    
    \pic [local bounding box=err3, right of=layer_3_3] {sum block={+}{}{-}{}};
    \draw[arrow] (layer_3_3) -- (err3);

    \node (target) [above of=err1, yshift=-1cm, font=\huge] {$T_i$};
    \draw[arrow] (target) -- (err1);
                                                    
    \draw[arrow] (err1) -- (err2);
                                                    
     \path (err2) -- (err3) node (errdots) [font=\Huge, midway, sloped] {$\dots$};
                                                    
     \draw[thick] (err2) -- (errdots);
     \draw[arrow] (errdots) -- (err3);

    \draw[arrow] (err1) edge[out=-90,in=330,red] (layer_1_1);

    \draw[arrow] (err2) edge[out=-90,in=330,red] (layer_2_2);

    \draw[arrow] (err3) edge[out=-90,in=330,red] (layer_3_3);

    \node (output)[right of=err2, yshift=-2cm,font=\huge] {$\Sigma$};
    \draw[arrow] (layer_1_1) edge[out=0,in=135,green] (output);
    \draw[arrow] (layer_2_2) edge[out=0,in=180,green] (output);
    \draw[arrow] (layer_3_3) edge[out=0,in=210,green] (output);
    \node (Y) [font=\huge, right of=output, xshift=-1cm] {$Y_i$};
    \draw[arrow] (output) -- (Y);
                                                    
    \node (output layer) [left of=target, xshift=-0cm, font=\Large, color=blue!80!white] {Output Layer};
    \node (hidden layers) [left of=output layer, xshift=-1.5cm,
      font=\Large, color=cyan] {Hidden Layers};

\end{tikzpicture} }

&
    \includegraphics[width=0.48\textwidth]{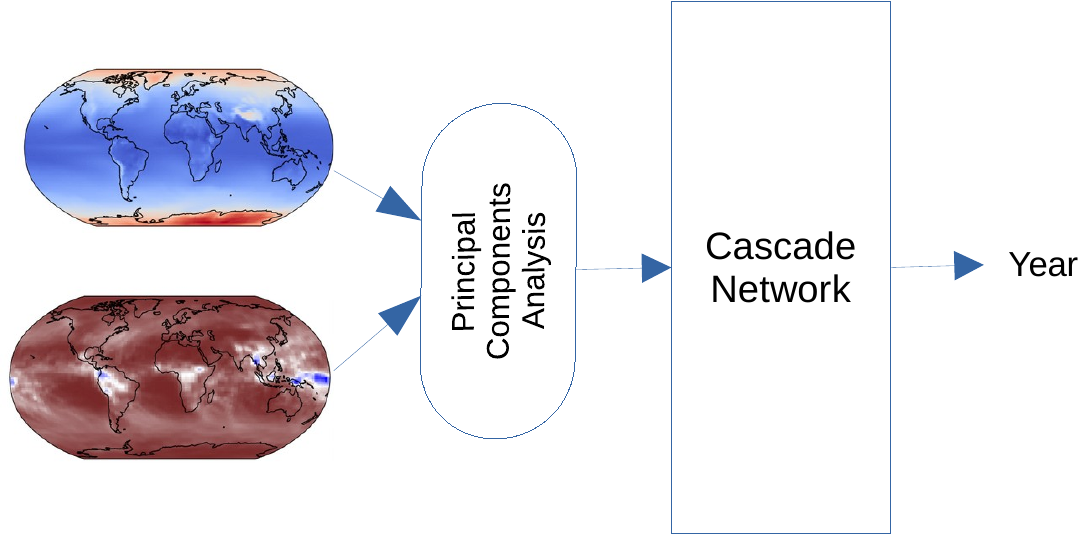}
\\
    a. & b.
    \end{tabular}
    \caption{a) Cascade Network.  b) Use for climate model data.}
    \label{fig:cascade-diagram}
\end{figure}
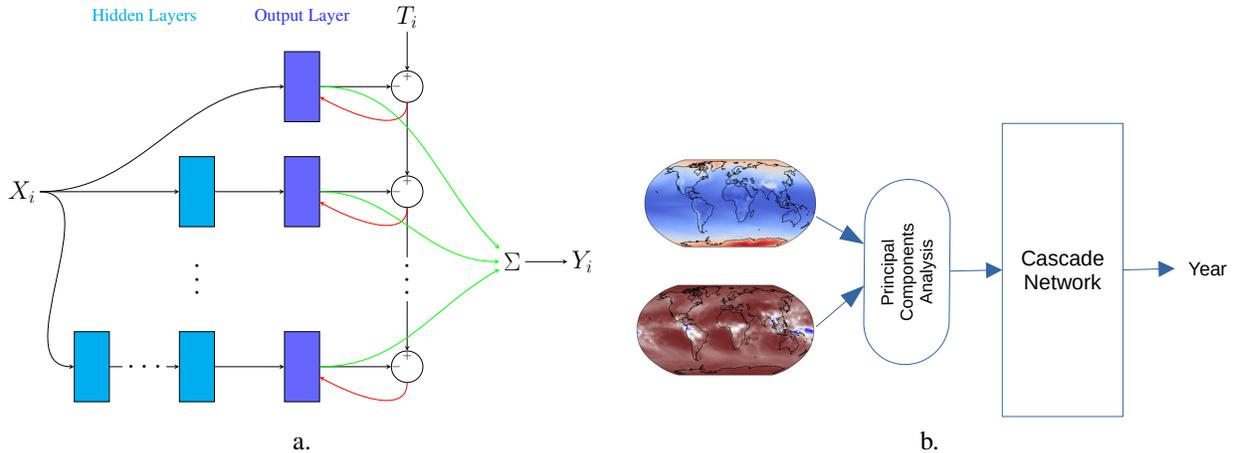

The resulting model structure can be viewed as an ensemble of neural networks
whose outputs are combined in a cascade structure, shown in
Figure~\ref{fig:cascade-diagram}a. For the regression problem studied in this
paper, the top network is a one-layer network of
linear units. This network is trained to minimize the mean squared
error in its output.  After this, a second network of one hidden layer
is trained to minimize the error in the first network's output.
This can also be described by defining the output of the second network to
be the sum of its output plus the output of the first network, leading
to the cascade structure.  This second network is only kept in the
model if the error on a validation set is reduced, otherwise it is
discarded.  This process of adding networks continues with networks of
more hidden layers.

\section{Models of Climate Change}

The sixth phase of the Coupled Model Intercomparison
Project~\cite{Eyring16} (CMIP6) has produced 35 models of earth's
atmosphere from which simulated global temperature and precipitation
maps can be obtained for years from 1850 to 2100.  This data can
be used to investigate spatial and temporal temperature and
precipitation patterns that relate to changes in climate.  For
example, Barnes, et al.~\cite{Barnes19,Barnes20}, used CMIP5 data to
map global temperature data to years. They found that linear models
predicted the year quite well.  They also recast the regression
problem as a classification problem in order to use Layerwise
Relevance Propagation (LRP) to identify spatial patterns significant
to classifying specific decades~\cite{Toms20}.

Here the regression approach is extended by including temperature and
precipitation CMIP6 data and modeling it with a cascade network. We
used this data with a spatial resolution of 120 latitudes and 240
longitudes. Not surprisingly, this data has numerous correlations
among spatial locations, which was dealt with in prior work by
performing ridge regression to limit the magnitude of weights in the
first layer of the neural network models~\cite{Barnes19}.  Here we use
an alternative approach---Principal Components Analysis (PCA) is used
to represent the data with independent factors and to decrease the
data dimensionality, illustrated in Figure~\ref{fig:cascade-diagram}b.

To reveal patterns in the data that relate to the predicted years, a
direct approach is used here.  Patterns in global temperature and
precipitation to which a trained network is most sensitive are
determined by reconstructing the original global maps from the 
the weights in the first layer of a network.

\section{Method}

Let $x_i$ be the $i^{th}$ input sample that is received by all
networks.  Let $y_{i,j} = f_j(x_i)$ be the output of the $j^{th}$ network for
input sample $x_i$.  For the regression problems considered here, 
the target value for the $i^{th}$ sample is the scalar $t_i$. As shown
in Figure~\ref{fig:cascade-diagram}a, the sum of the outputs,
$Y_i = \sum_{j=1}^M y_{i,j}$  forms the
output of the cascaded nets.  The objective is to minimize the sum of
squared errors $\sum_{i=1}^N (t_i - Y_i)^2$.

The optimization problem is structured to encourage the simpler nets,
higher in the cascade shown in Figure~\ref{fig:cascade-diagram}, to
reduce the error as much as possible before recruiting a more complex
network to reduce the error further.  To accomplish this, the target
value for net $j$ is the original target minus the sum of the outputs
of the previous, simpler networks: the target value, $t'_{i,j}$, for input
$i$ for net $j$ is $t'_{i,j} = t_i - \sum_{k=1}^{j-1} y_{i,j}$.

The CMIP6 data consists of simulated annual temperature and precipitation
from 35 models for 251 years from 1850 through 2100, at 120 latitudes and 240
longitudes. The mean annual temperatures and precipitations across the
globe are removed from each sample. This data is divided into
training, validation, and testing partitions selected from different
years. The training partition was assigned 50\% of the data, and the
validation and testing partitions were each assigned remaining 25\% of
the data.  Each partition contained data from all 35 atmosphere
models, but for different sets of years.  The partitioning can be seen
in Figures~\ref{fig:err-vs-pcs} and \ref{fig:err-vs-nets}.

Optimization of the networks is performed on the training partition
using the Scaled Conjugate Gradient algorithm~\cite{Moller93}.  PCA is
performed by projecting the flattened and concatenated temperature and precipitation
maps onto 1 to 500 principal components.  The structures of the neural
networks in the cascade structure have increasing numbers of hidden
layers, each layer containing only two units having the hyperbolic
tangent activation functions. This number of units was
found to generalize better to validation and testing data than larger
numbers of units.

\section{Results}

The results of the training procedure are shown in
the following figures. Figures~\ref{fig:err-vs-pcs}a and b plot the
RMSE in predicted year with respect to the number of principal
components (PCs).  Each of our data samples contain $120\times 240$
values for temperature and also for precipitation, resulting in
57,600 values. During our training experiments, we tested the use of
1 to 500 PCs, a large reduction in dimensionality of the data.  More
PCs resulted in higher generalization error.

Figure~\ref{fig:err-vs-pcs}a shows that RMSE for the three data
partitions are similar, with the best number of PCs, determined by the
lowest validation error, was found to be 240.
Figure~\ref{fig:err-vs-pcs}b is a plot of the number of neural nets
included in the cascade nets for each number of PCs.  With a small
number of PCs, the cascade keeps up to 7 nets, but as the number of
PCs is increased, fewer nets are needed to maintain a low
error. Recall that nets increase in complexity (number of hidden
layers) but are kept in the cascade only if their inclusion decreases
the validation error.  The cascade trained with 240 PCs contains two
neural nets, one being the linear net, and one with a single hidden
layer.

Figure~\ref{fig:err-vs-pcs}c shows the RMSE in predicted year for the
best cascade, the one with 240 PCs.  The error for a linear network,
with 0 hidden layers, 27 to 30 years.  When the single hidden layer net is
added to the cascade the error is reduced to 20 to 22
years. Figure~\ref{fig:err-vs-pcs}d shows the predicted year versus
the actual year.  Perfect prediction would result in all points
falling along the red diagonal line.  The partitioning of the data
into training, validation, and testing subsets is clear in this plot.
It is clear that data from later years, from 2000 to 2100 is predicted
better by the cascade.  Is is also apparent that data for the years
from 1850 to about 1950 is harder to predict.  Perhaps this is due to
a lack of information in the temperature and precipitation maps that
are related to specific years in this early period.


\begin{figure}[!ht]
  \centering
  \begin{tabular}{cc}
    \includegraphics[width=0.5\textwidth]{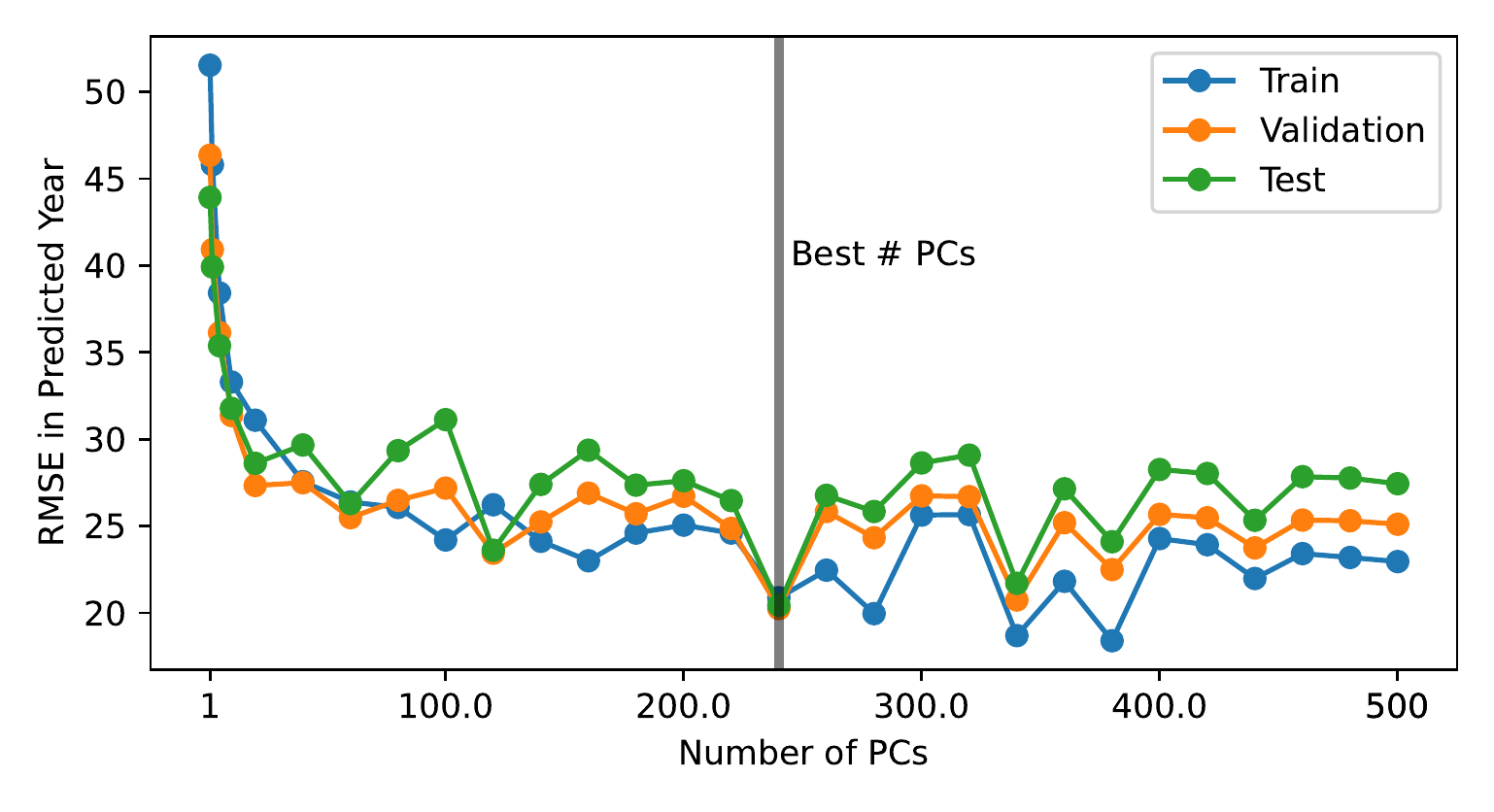}  &
    \includegraphics[width=0.5\textwidth]{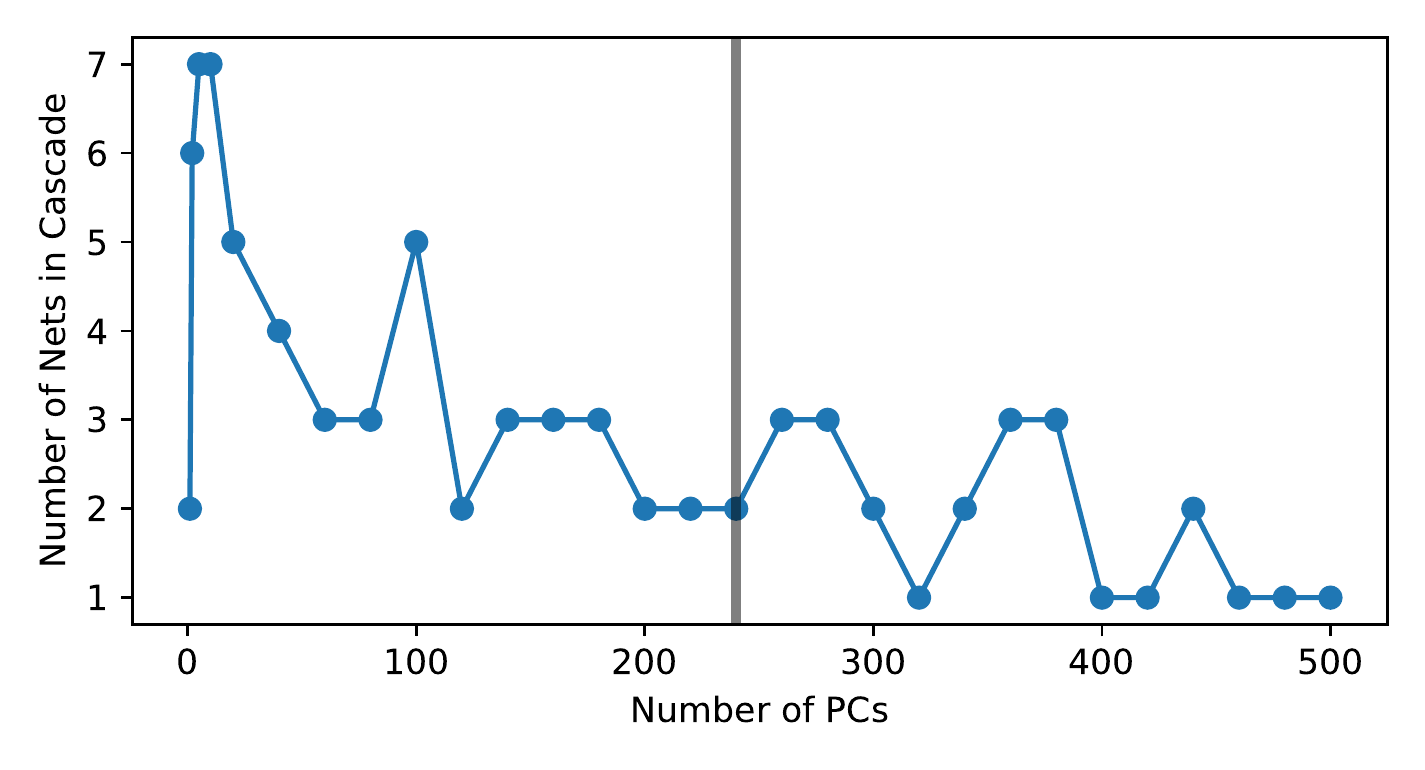} \\
    a. & b. \\
    \includegraphics[width=0.45\textwidth]{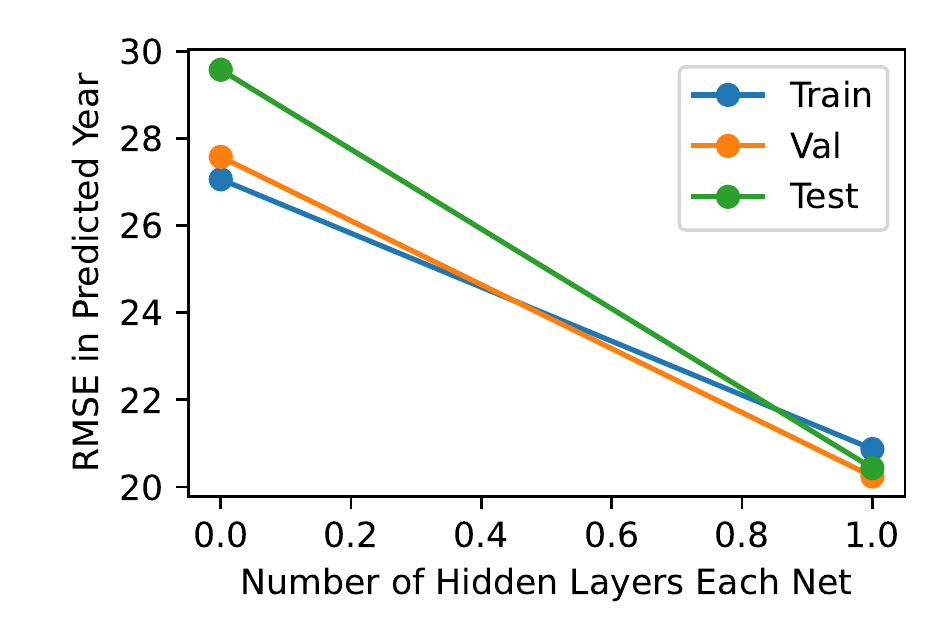}  &
    \includegraphics[width=0.45\textwidth]{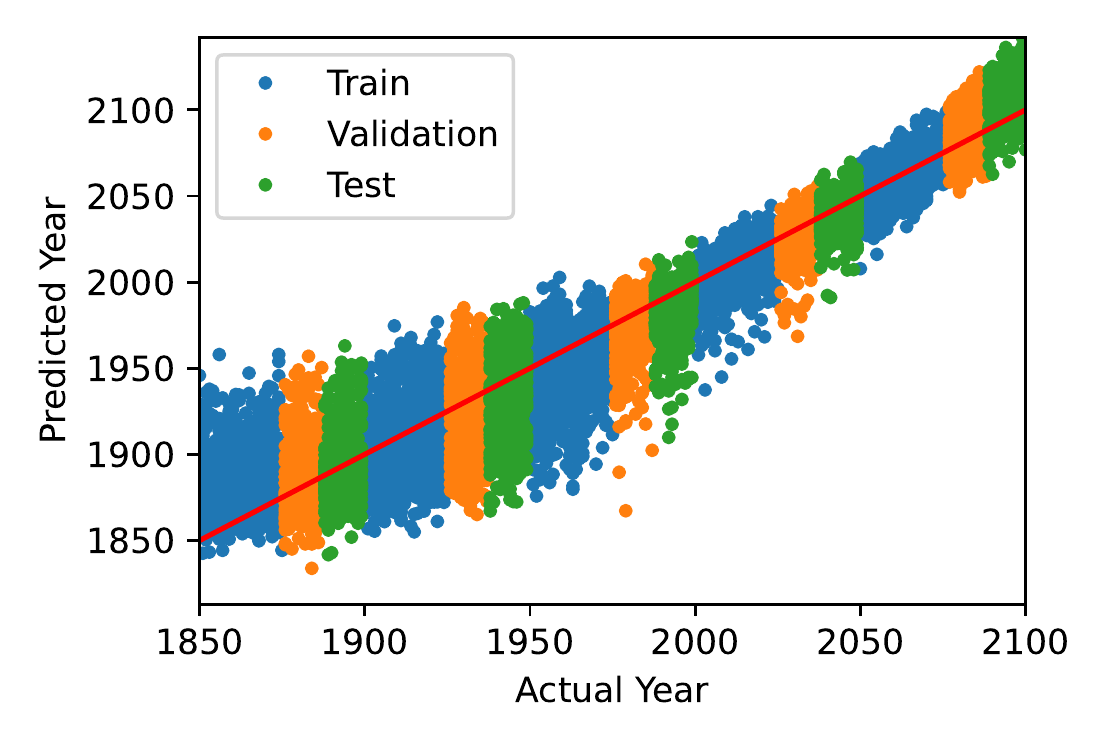}\\
    c. & d.
  \end{tabular}
    \caption{ a) RMSE in predicted year versus the number of principal
    components (PCs). The best number of PCs at 240 is determined by
    the lowest validation RMSE.  b) Number of nets in the cascade
    model versus number of PCs. c) RMSE in predicted years for the
    best cascade model due to each of the two nets. d) Predicted year
    versus the actual year for all samples.}
    \label{fig:err-vs-pcs}
\end{figure}

\begin{figure}[!ht]
  \centering
  \begin{tabular}{cc}
    \multicolumn{2}{c}{\includegraphics[width=0.7\textwidth]{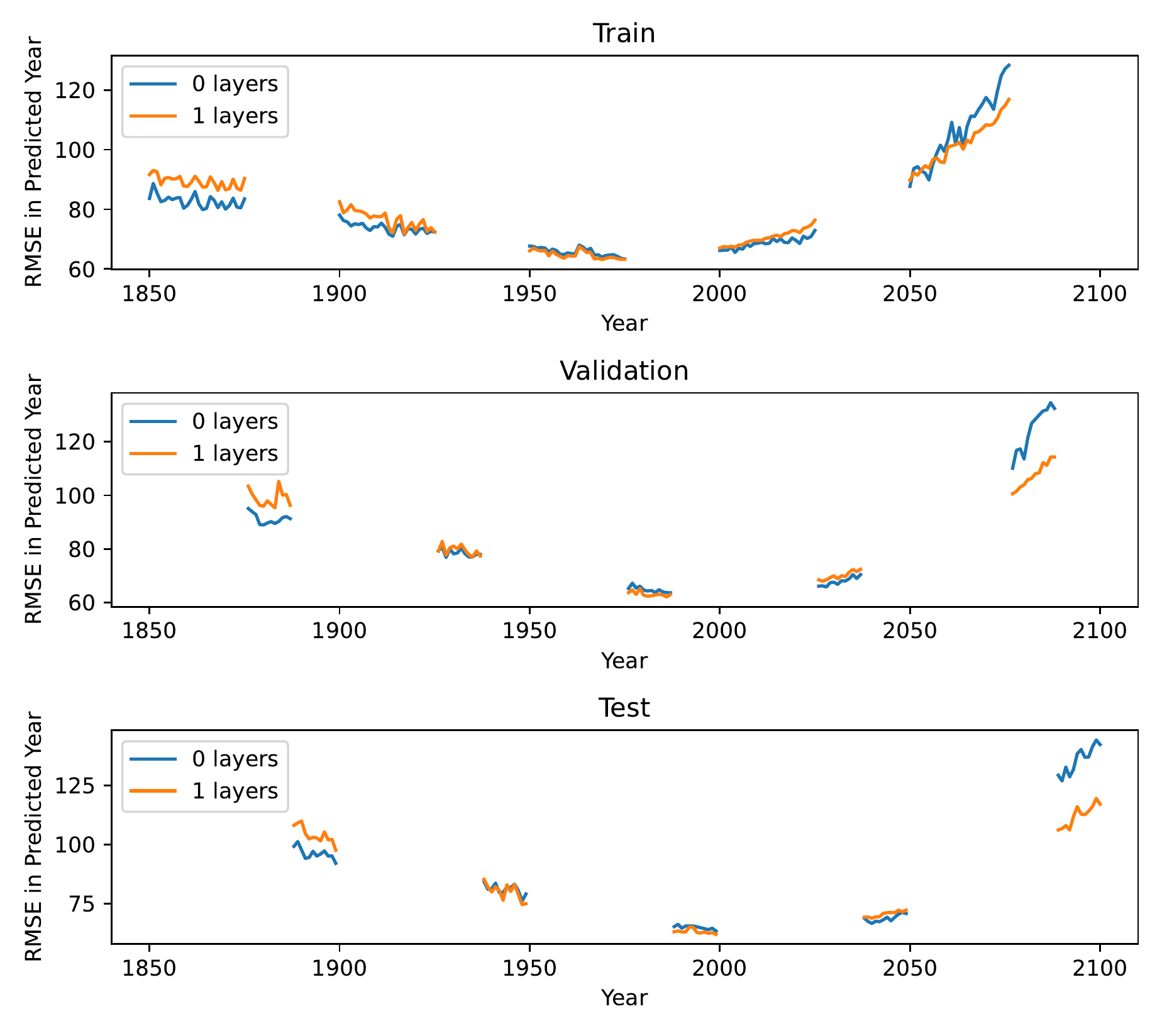}}\\
  \end{tabular}
    \caption{RMSE in predicted year using just the linear network and
      with the cascade model output that is the combination of the
      linear network and the one-hidden layer network.}
    \label{fig:err-vs-nets}
\end{figure}

Since the cascade model consists of two networks, one with no hidden
layers and one with one hidden layer, an obvious question is whether
or not there is a span of years with the second network most helps in
predicting the year.  Figure~\ref{fig:err-vs-nets} shows that the
answer is ``yes''.  This figure shows the RMSE in predicted year
versus the actual year for the three data partitions.  In each plot
the RMSE is shown for just the linear network, then again for the
combination of the linear and the nonlinear network with a single
hidden layer. The second and third plots, for the validation and test
partitions, respectively, show that the addition of the second
network with one hidden layer with nonlinear activation functions improves the prediction of the year (decreases the error) for
the latest span of years.  This suggests that the patterns in the data
that the second network have acquired may indicate aspects of the
global temperature and precipitation data that distinguish early from
late years, and thus help identify climate change indicators.

\begin{figure}[!ht]
  \centering
  \begin{tabular}{c}
    \includegraphics[width=0.5\textwidth]{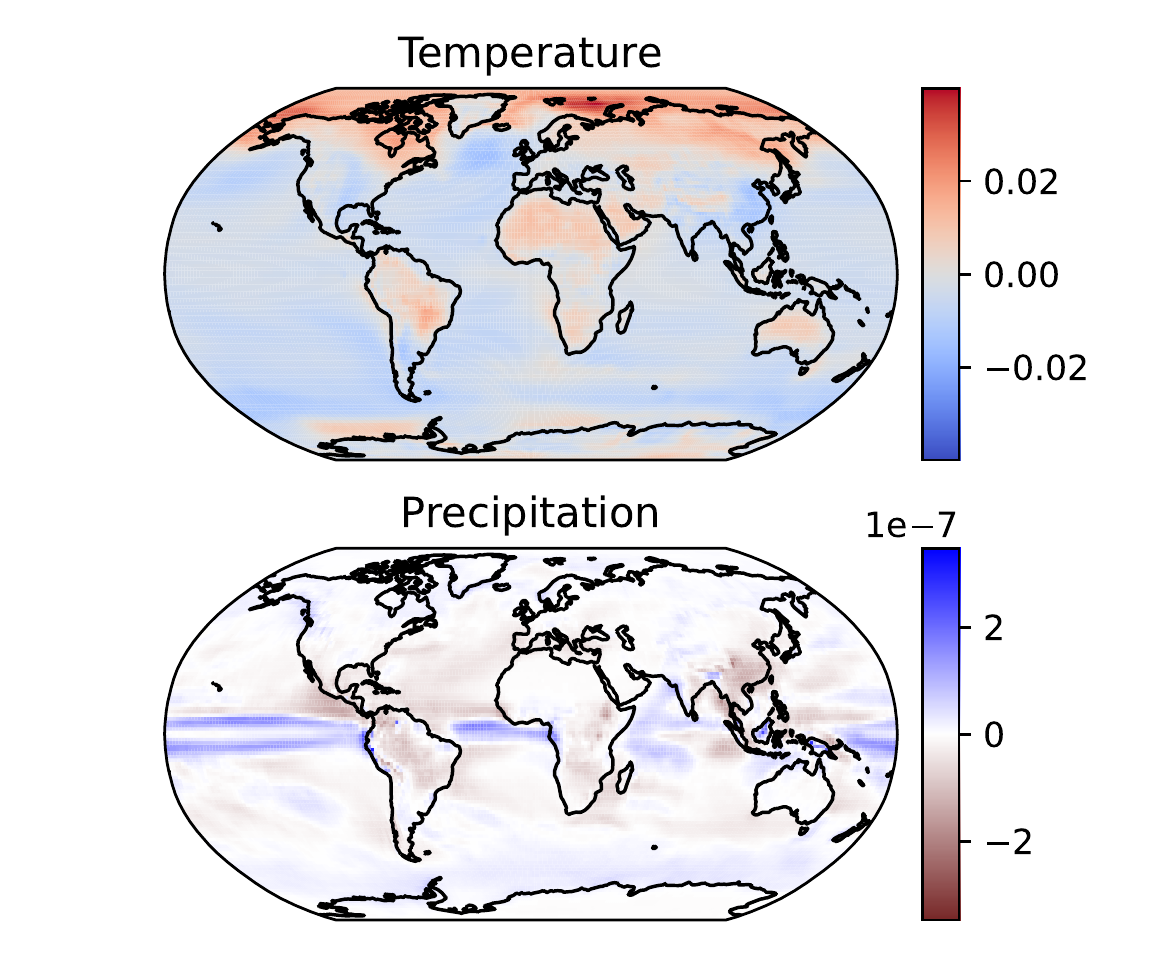}\\
    a.\\
    \includegraphics[width=0.9\textwidth]{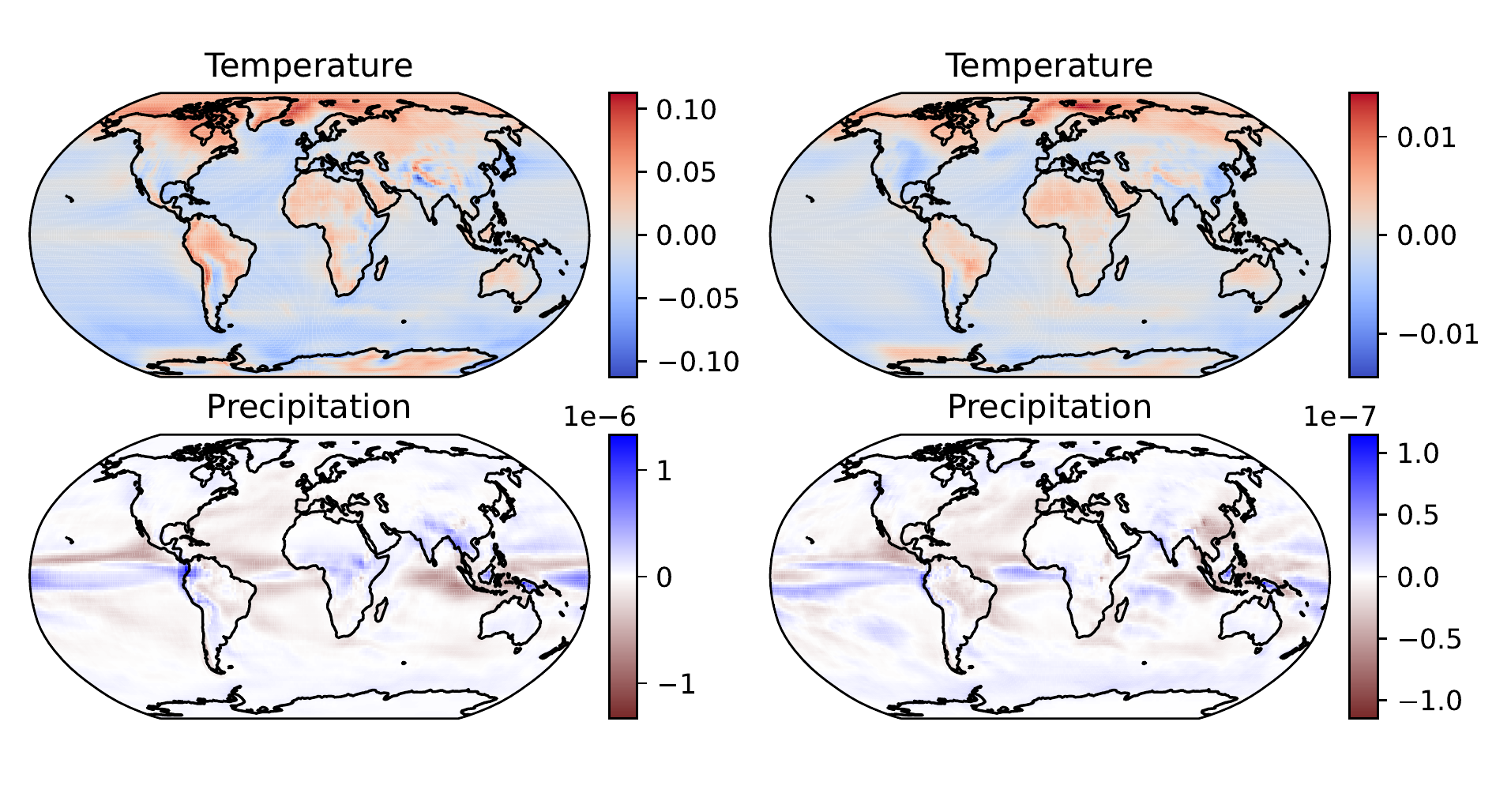} \\
     b.
  \end{tabular}
    \caption{ Temperature and precipitation maps related to
    year. a) Maps from the first linear network.  b. Maps from the two
    units in the single hidden layer of the second network.}
    \label{fig:maps}
\end{figure}

To visualize these patterns, the following procedure was used. The
weights in the units of the first layer of a network were used to
reconstruct the global temperature and precipitation maps by combining
each principal component multiplied by the corresponding
weights. Figure~\ref{fig:maps}a shows the result for the linear
network. The temperature map is not surprising; indicates that higher
temperatures in the arctic are positively related to an increase in
year. A closer examination shows that increased temperature in the
Barents Sea north of Finland may be more strongly positively related to year.
The precipitation map reveals a higher significance in
precipitation near the equator as a positive relationship to year.

Figure~\ref{fig:maps}b shows the maps from the two units in the first,
and only, hidden layer of the second network.  Differences among the
temperature maps are not clear, though can be revealed with
higher-resolution images. More obvious differences exist in the
precipitation maps. It appears that these maps show an 
asymmetry above and below the equator in positive and negative
relationships with year over the Pacific ocean.  Another asymmetry
appears between precipitation in the Pacific versus the Atlantic
oceans.  These asymmetries are not apparent in the precipitation map
for the linear network. Since the addition
of this network lowered the error for the latest years, these maps may
reveal key differences in precipitation in late versus early years in
the 1850-2100 span.

\section{Conclusion}

An ensemble of neural networks is developed in a cascade structure in
which simpler networks are trained first and more complex ones, with
additional hidden layers, are trained and kept only if they result in
lower error on a validation set.  The result is a model  that can be 
interpreted by assembling aspects of the data to which each network is
learned to be sensitive.

The training of a cascade model of temperature and precipitation changes with respect
to year results in correct year prediction with an approximate error
of 20 years over the span of 1850 to 2100.  The data only supports a
single-hidden layer net added to the initial linear net.  Additional,
more complex nets increase the validation error. Similar increases in
validation error occur with larger numbers of principal components
used to project the 57,600 to smaller dimensions.

Current experiments with the cascade structure are underway with data
sets that contain stronger nonlinearities, and with classification problems.

\subsubsection*{Acknowledgments}
This work was partially funded by NSF Grant \#2019758, \textit{AI Institute: Artificial Intelligence for Environmental Sciences (AI2ES)}

\bibliography{cascadenet}
\bibliographystyle{plain}

\end{document}